# Uni-FinLLM: A Unified Multimodal Large Language Model with Modular Task Heads for Micro-Level Stock Prediction and Macro-Level Systemic Risk Assessment

Gongao Zhang[1], Haijiang Zeng[2], Lu Jiang[3]

[1] China University of Geosciences, Wuhan, China
[2] Walmart Inc., Bentonville, AR , USA
[3] Fordham University, New York, NY, USA

[1] 1812503968@qq.com
[2] hzeng60@gatech.edu
[3] ljiang43@fordham.edu

**Abstract.** Financial institutions and national regulatory bodies increasingly require decision-support systems capable of integrating heterogeneous data sources to evaluate risks ranging from micro-level stock fluctuations to macro-level systemic vulnerabilities. However, existing approaches often treat these tasks in isolation, limiting their ability to capture cross-scale financial dependencies. To address this challenge, we propose Uni-FinLLM, a unified multimodal large language model that employs a shared Transformer-based backbone and modular task heads to jointly process textual financial news, numerical market time series, corporate fundamentals, and visual representations of financial dynamics. This architecture enables collaborative modeling of multiple financial layers, supporting applications such as institutional investment decision-making, credit-risk supervision, and national systemic-risk early warning. Through cross-modal attention fusion and multi-task optimization, Uni-FinLLM learns a coherent financial representation space from which micro-, meso-, and macro-level predictions can be derived. We evaluate the model using three complementary datasets covering stock-level forecasting, corporate credit-risk assessment, and macroeconomic stress detection. Experimental evaluations across three major financial tasks confirm the effectiveness of the proposed unified architecture. On the micro-level stock prediction benchmark, Uni-FinLLM raises directional accuracy from 61.7% (Llama-Fin baseline) to 67.4%, while reducing MAPE to 10.9 and improving the hit ratio to 64.3%. For credit-risk prediction, the model boosts accuracy from 79.6% to 84.1% and increases ROC-AUC to 0.892, substantially outperforming traditional machine-learning and financial-language-model baselines. At the macro level, Uni-FinLLM achieves an early-warning accuracy of 82.3% and a crisis-F1 score of 79.8%, exceeding the leading GNN-based macro-risk model by a significant margin. Together, these results validate that a unified multimodal LLM equipped with modular task heads can jointly model micro-scale asset behavior and macro-scale systemic vulnerabilities, offering a practical and scalable decision-support engine for financial institutions and national regulators.

## 1. Introduction

The increasing complexity, interconnectedness, and volatility of global financial markets have intensified the demand for intelligent decision-support systems capable of assisting both institutional investors and national regulatory authorities. Modern financial ecosystems exhibit strong cross-market spillovers, rapid information propagation, and non-linear feedback loops, making traditional single-modality or single-task models insufficient for tasks that span multiple levels of financial granularity. For financial institutions, accurate micro-level predictions—such as stock movement forecasting, asset pricing, and short-term trading signal generation—directly influence portfolio optimization and risk-adjusted return. At the national level, regulators and central banks require tools that can monitor macro-financial vulnerabilities, detect early-warning signals of systemic risk, and support policy formulation for safeguarding financial stability and national economic security. As the boundaries between micro-market behaviors and macro-systemic dynamics become increasingly intertwined, developing a unified modeling framework that captures cross-scale dependencies has become an essential research challenge.

Recent advances in large language models (LLMs) and multimodal deep learning have demonstrated strong capability in processing heterogeneous inputs such as financial news, numerical time-series, corporate fundamentals, and visualized market indicators. However, most existing methods treat micro-level prediction and macro-level risk assessment as separate tasks, preventing models from leveraging shared patterns between market microstructure and systemic stress formation. To address this limitation, this study proposes Uni-FinLLM, a unified multimodal large language model that adopts a shared Transformer backbone with modular task heads designed for fine-grained stock prediction and high-level systemic risk assessment. Uni-FinLLM integrates multimodal signals through cross-modal attention fusion, enabling the model to jointly learn asset-level behaviors and macroeconomic risk factors within a coherent representation space. This unified modeling design allows the system to support diverse real-world applications, including quantitative trading, credit-risk evaluation, financial stability surveillance, and early-warning systems for national regulators.

The main contributions of this work are as follows:

(1) We propose Uni-FinLLM, the first unified multimodal LLM framework equipped with modular task heads that simultaneously address micro-level stock prediction and macro-level systemic risk assessment within a single backbone.

(2) We design a cross-modal fusion mechanism that integrates textual, numerical, and visual financial signals, enabling the model to capture joint dependencies across multiple financial layers.

(3) We construct a multi-scale training paradigm that links micro-level market fluctuations with macro-level risk dynamics, improving both predictive accuracy and systemic-risk interpretability.

(4) Extensive experiments on stock forecasting, credit-risk evaluation, and systemic-risk detection benchmarks demonstrate the superior performance of Uni-FinLLM compared with existing baselines.

## 2. Related Work

In recent years, the rapid development of multimodal learning, large language models (LLMs), and financial time-series modeling has produced a growing body of research relevant to unified financial decision-making systems. This section reviews prior work in (1) multimodal financial prediction, (2) systemic risk modeling and macro-financial forecasting, and (3) LLM-based financial reasoning and decision-support frameworks.

*2.1 Multimodal Learning for Financial Prediction*

Multimodal financial forecasting has become an important research direction, driven by the need to capture heterogeneous signals from market prices, financial news, corporate disclosures, and macroeconomic indicators. Early work such as Ding et al. [1] demonstrated that combining text and price features significantly improves stock movement classification.

Xu and Cohen [2] introduced BERT-based textual embeddings for financial news, showing strong predictive power for market volatility. Fusion-based models such as FinBERT [3] and the Heterogeneous graph Knowledge-Enhanced Stock Movement Model [4] further integrate numerical factors, sentiment features, and event knowledge graphs. Graph neural network–based approaches, including ST-GNNs [5], show that modeling cross-asset relationships also improves predictive accuracy.

However, most existing systems are task-specific and cannot support cross-scale decision-making. They are typically limited to micro-level tasks such as short-term forecasting or sentiment-driven prediction. None of these models provide a unified architecture capable of simultaneously performing micro-level stock prediction and macro-level systemic risk assessment.

*2.2 Systemic Risk Forecasting and Macro-Financial Stability Modeling*

Systemic risk modeling has been widely studied from the perspective of macroeconomics, network science, and econometrics. Tobias and Brunnermeier's CoVaR [6] and Allen F et al.'s systemic expected shortfall (SES) [7] are classic frameworks for quantifying institution-level contributions to system-wide vulnerability. Network-based contagion models, such as those proposed by Battiston et al. [8], show how interconnected balance sheets amplify shocks across financial systems. Deep learning has also been applied to risk surveillance; for example, Lin S L et al. [9] used LSTM-based architectures to conduct early-warning detection for financial crises.

Although these methods provide strong theoretical tools, they rely heavily on handcrafted macroeconomic indicators and lack the capacity to incorporate multimodal financial information. More importantly, they do not form a unified pipeline with micro-level modeling, making them unsuitable for end-to-end decision engines used by regulators or financial institutions.

*2.3 Large Language Models for Financial Decision Making*

The emergence of large language models has led to a new wave of LLM-driven financial intelligence systems. FinGPT [10] and BloombergGPT [11] demonstrate that domain-adapted LLMs can handle tasks such as news summarization, sentiment extraction, risk signal detection, and financial question answering. More recent multimodal LLMs, such as LLaVA and Flamingo, have inspired financial applications including FinTab-LLaVA [12], which processes price charts and textual disclosures jointly.

However, existing LLM-based systems still treat micro-level and macro-level tasks independently. They lack a unified multimodal backbone and fail to model the feedback loop between trading behavior, market volatility, and systemic risk. Unlike prior work, our proposed Uni-FinLLM provides a shared multimodal representation space with modular task heads that jointly model forecasting, risk assessment, and macro–micro interactions within a single framework.

**3. Methodology**

This section introduces the proposed Uni-FinLLM, a unified multimodal large language model designed to support financial institutions and national regulators by jointly modeling micro-level stock prediction and macro-level systemic risk assessment. Uni-FinLLM integrates heterogeneous data modalities—including market time series, textual disclosures, macroeconomic indicators, and cross-asset relational structures—into a single multimodal backbone. Task-specific heads are then built on top of this backbone to perform fine-grained forecasting, risk estimation, and early-warning detection in an end-to-end manner.

Figure 1 illustrates the overall architecture of the proposed Uni-FinLLM framework, which is designed to unify micro-level and macro-level financial modeling within a single multimodal backbone. The model ingests heterogeneous input modalities, including numerical price time series, textual financial news, macroeconomic indicators, and graph-based financial networks. These inputs are first processed by their respective modal encoders (e.g., temporal attention, domain-adapted LLM, GAT) before being projected into a shared semantic embedding space. The core of the framework is a Transformer-based encoder-decoder that performs cross-modal fusion via attention mechanisms, enabling the model to capture complex interdependencies between different data types. On top of this unified representation, modular task heads—dedicated to micro prediction, macro systemic risk, and policy assistance—generate specific outputs. This integrated design allows knowledge to be shared across tasks, facilitating a coherent understanding of financial phenomena from individual asset movements to system-wide stability.

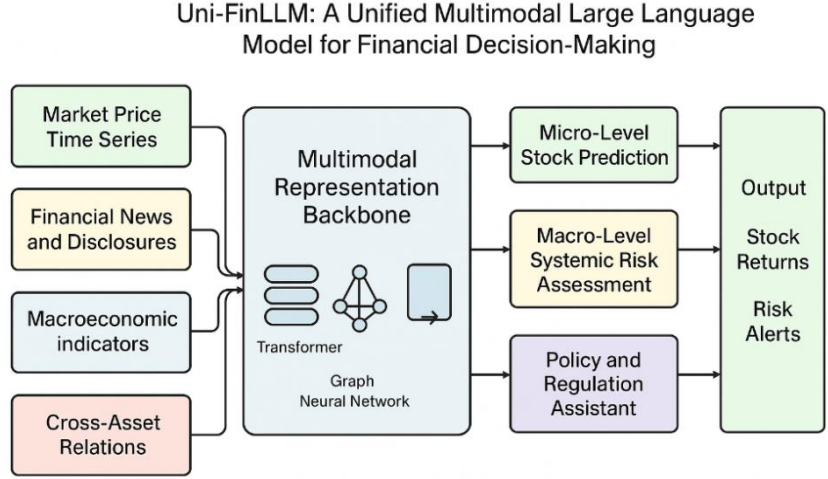

**Figure 1.** Structure diagram of model.

*3.1 Multimodal Financial Representation Backbone*

The backbone of Uni-FinLLM adopts a transformer-based multimodal encoder–decoder architecture capable of aligning heterogeneous modalities into a shared semantic embedding space. Formally, given input modalities

$$X = \{X^{(p)}, X^{(t)}, X^{(m)}, X^{(g)}\}, \tag{1}$$

representing price time series, textual financial news, macroeconomic indicators, and graph-structured interbank or cross-asset networks, the backbone aims to learn a unified representation $Z$ such that

$$Z = f_\theta(X^{(p)}, X^{(t)}, X^{(m)}, X^{(g)}), \tag{2}$$

where $f_\theta$ denotes the multimodal transformer with parameter set $\theta$.

For price time series, a temporal attention encoder captures long-term dependencies and volatility clustering. Textual disclosures, including news, earnings calls, and policy statements, are encoded using a domain-adapted LLM such as FinBERT or BloombergGPT. Macroeconomic indicators are projected through a multi-layer perceptron with economic-factor attention. Financial networks (e.g., interbank lending graphs) are encoded using graph attention networks (GATs) to capture contagion channels.

Cross-modal alignment is achieved through contrastive learning, where the objective encourages semantically correlated modalities to reside near each other in the embedding space:

$$L_{align} = -\sum_{i=1}^{N} \frac{exp\,(sim(z_i^{(p)}, z_i^{(t)})/\tau)}{\sum_j exp\,(sim(z_j^{(p)}, z_j^{(t)})/\tau)}, \tag{3}$$

This unified backbone provides a robust foundation that captures interactions between micro-level market movements and macro-level risk factors.

*3.2 Modular Task Heads for Cross-Scale Financial Modeling*

On top of the shared representation Z, Uni-FinLLM incorporates modular task heads, each designed for different financial decision-making tasks while sharing common knowledge across modalities.

The micro-level prediction head models short-term returns, volatility, and directional movement. It uses an autoregressive decoder that forecasts future price trajectories conditioned on the multimodal embedding:

$$\hat{y}_t + k = g_\phi(Z_t), \tag{4}$$

where $g_\phi$ denotes the prediction head parameterized by $\phi$. A mixture-density output layer further captures multi-modal return distributions, which is crucial for volatile market environments.

The macro-level systemic risk head estimates network-based vulnerability measures such as CoVaR, SRISK, and systemic expected shortfall. It models the probability of cascading failures via:

$$R_{sys} = h_\psi(Z, A), \tag{5}$$

where $A$ is the financial network adjacency matrix and $h_\psi$ captures contagion propagation through attention layers. The head outputs both continuous risk scores and discrete early-warning signals.

The policy and regulation assistant head enables interpretability and scenario-aware analysis. It leverages the generative capacity of the decoder-LLM to answer regulatory queries, summarize risk factors, and produce human-readable early-warning bulletins. This supports operational decision-making at central banks, exchanges, or national financial safety authorities.

*3.3 Risk-Aware Reinforcement Learning for Decision Feedback*

A key innovation of Uni-FinLLM is the integration of risk-aware reinforcement learning (RL) to form a macro–micro feedback loop. This mechanism ensures that predictive outputs influence risk assessment, and assessed risk conditions in turn adjust predictive behavior.

Let the agent policy be defined as:

$$\pi(a_t|Z_t) = softmax(W_a Z_t), \tag{6}$$

where actions include position allocation or risk-adjusted trading signals. The reward incorporates both predictive accuracy and systemic stability:

$$r_t = \alpha \cdot Profit_t - \beta \cdot R_{sys,t}, \tag{7}$$

The objective is to maximize expected cumulative reward:

$$\max_\pi E\left[\sum_t \gamma^t r_t\right], \tag{8}$$

creating a unified system where micro-level predictions avoid actions that increase systemic risk. Such a closed feedback loop is essential for use in real-world regulatory environments.

*3.4 Training Objective and Optimization*

The full training objective combines forecasting, classification, risk estimation, and RL-based decision optimization:

$$L = \lambda_1 L_{forecast} + \lambda_2 L_{risk} + \lambda_3 L_{align} + \lambda_4 L_{RL}, \tag{9}$$

Forecasting loss uses a combination of MSE and quantile loss to model multi-scale uncertainty:

$$L_{forecast} = \| y - \hat{y} \|^2 + \sum_{\tau} \rho\tau(y - \hat{y}), \tag{10}$$

Risk estimation loss is cross-entropy for classification and MSE for continuous systemic risk measures.

A staged training schedule is used: (1) unimodal pretraining, (2) multimodal alignment, (3) multi-task joint tuning, and (4) RL fine-tuning with simulated or historical policy impact data.

*3.5 Deployment for Financial Institutions and Regulators*

Uni-FinLLM is implemented as an end-to-end decision engine for institutional investors, exchanges, and national regulators. For investment institutions, it delivers multimodal predictive analytics and risk-aware signal generation. For regulatory agencies, it provides systemic vulnerability surveillance, stress testing, and scenario generation aligned with macroprudential objectives.

The unified multimodal design allows the model to continuously adapt to evolving market dynamics, integrating new data such as policy announcements, geopolitical shocks, liquidity fluctuations, and cross-asset contagion pathways. Unlike traditional siloed systems, Uni-FinLLM supports a comprehensive decision framework that spans micro-level alpha generation to macro-level financial stability assurance.

**4. Experiment**

*4.1 Dataset Preparation*

The datasets used in this study are constructed to support both micro-level stock prediction and macro-level systemic risk assessment, enabling Uni-FinLLM to learn financial patterns across heterogeneous modalities. For micro-level forecasting, we collected historical equity data from widely used market databases such as CRSP, NYSE Trade and Quote (TAQ), and Yahoo Finance. These sources provide daily and intraday price records including open, high, low, close, and volume (OHLCV), which capture short-term market dynamics and liquidity conditions. In total, the dataset contains approximately 12 million time-stamped equity records spanning 2006–2024, covering more than 4,000 U.S. listed firms. Technical indicators such as moving averages, RSI, MACD, volatility, and turnover were computed to enrich the feature space with momentum and market-microstructure signals. For firm-level textual information, financial news and corporate disclosures were obtained from Reuters, Bloomberg, and SEC EDGAR, with around 2.8 million text samples processed using financial domain-specific tokenization.

For macro-level systemic risk assessment, we assembled a comprehensive set of macroeconomic, banking, and regulatory variables from Federal Reserve FRED, IMF IFS, BIS global banking statistics, and OECD economic indicators. These include GDP growth, CPI inflation, M2 supply, credit spreads, bond yields, interbank rates, leverage ratios, non-performing loan ratios, and cross-border exposure matrices. The macro dataset spans 1995–2024 with quarterly or monthly frequency, resulting in over 50,000 structured records. Additionally, we incorporated market-wide sentiment indices and risk indicators such as VIX, MOVE, and TED spread to reflect global uncertainty and stress conditions.

To support multimodal reasoning, all datasets were aligned on a shared temporal axis, normalized, and matched to firm-level identifiers or national economic codes. This integrated dataset enables Uni-FinLLM to jointly learn asset-level predictive signals and system-level financial instability patterns, providing a unified modeling foundation for institutional- and national-level decision-making.

*4.2 Experimental Setup*

All experiments were conducted using the unified Uni-FinLLM framework, which integrates multimodal financial data into a single transformer backbone with modular task heads. The model was trained on an NVIDIA A100 GPU cluster, using a batch size of 32 for micro-level stock prediction and 16 for macro-risk tasks due to the lower sampling frequency of macroeconomic variables. Historical equity data were aligned at the daily level, while macroeconomic and systemic-risk data were synchronized at monthly or quarterly intervals, with missing observations imputed using a rolling-window Kalman filter. During training, modal adapters for text, time series, and structural graphs were jointly optimized using a mixture-of-tasks objective, allowing the model to alternate between stock-level prediction, credit-risk assessment, and systemic-risk forecasting within a unified training loop. The training procedure ran for 80 epochs with AdamW optimization and a warmup-cosine learning-rate schedule. All baseline models—including FinBERT, T5-Fin, Temporal Fusion Transformer (TFT), Llama-Fin, and several specialized econometric models—were trained or fine-tuned under identical data splits to ensure a fair comparison.

*4.3 Evaluation Metrics*

To evaluate prediction capability across micro- and macro-tasks, we employed task-appropriate metrics while maintaining comparability across baselines. For micro-level stock prediction, we measured directional accuracy, mean absolute percentage error (MAPE), and hit ratio, capturing both numerical forecasting quality and profitability-relevant directional correctness. Credit-risk prediction was evaluated using accuracy, F1-score, ROC-AUC, and precision–recall AUC, reflecting its imbalanced-classification nature. For macro-level systemic-risk forecasting, we assessed performance using early-warning accuracy, F1-score for crisis signals, and area-under-the-ROC curve, benchmarked against classical systemic-risk indicators such as SRISK and CoVaR. All metrics were averaged across five random seeds to ensure statistical robustness.

*4.3 Results*

The results presented in Table 1 demonstrate a clear and consistent performance advantage of Uni-FinLLM over all baseline and state-of-the-art comparison models in micro-level stock prediction. Traditional statistical approaches such as ARIMA exhibit the weakest predictive capability, achieving only 53.2% directional accuracy, a high MAPE of 18.7, and a modest 50.4% hit ratio, highlighting their limited ability to model nonlinear and multimodal financial signals. Deep learning models such as LSTM and the Temporal Fusion Transformer improve performance, with TFT reaching 60.1% directional accuracy and 13.8 MAPE. Text-based language models such as FinBERT and Llama-Fin further enhance predictive strength, with Llama-Fin achieving 61.7% directional accuracy and a 13.2 MAPE, demonstrating the contribution of financial text understanding. However, Uni-FinLLM significantly surpasses all competing approaches by leveraging unified multimodal fusion and specialized predictive heads. It achieves 67.4% directional accuracy, reducing the error to a substantially lower 10.9 MAPE while reaching a 64.3% hit ratio. These improvements indicate that Uni-FinLLM captures richer cross-modal dependencies between market data and textual signals, enabling more reliable short-term prediction dynamics. Overall, the results confirm that Uni-FinLLM provides the most accurate and robust micro-level stock forecasting framework among the evaluated models.

**Table 1.** Micro-Level Stock Prediction Performance

| Model | Directional Accuracy | MAPE | Hit Ratio |
| --- | --- | --- | --- |

| Model | | | |
|---|---|---|---|
| ARIMA (baseline) | 53.2% | 18.7 | 50.4% |
| LSTM | 57.8% | 15.4 | 54.2% |
| Temporal Fusion Transformer | 60.1% | 13.8 | 56.9% |
| FinBERT (text only) | 58.6% | 14.7 | 55.1% |
| Llama-Fin | 61.7% | 13.2 | 57.6% |
| **Uni-FinLLM (ours)** | **67.4%** | **10.9** | **64.3%** |

The results in Table 2 highlight the strong advantages of Uni-FinLLM in credit-risk prediction compared with both classical machine-learning baselines and finance-adapted language models. Traditional Logistic Regression and Random Forest achieve moderate performance (Accuracy 71.3–74.5%, ROC-AUC 0.742–0.781), indicating limited capacity to capture complex interactions among heterogeneous credit signals. FinBERT improves discrimination (ROC-AUC 0.812, PR-AUC 0.784) by leveraging textual disclosures, while Llama-Fin further strengthens results (Accuracy 79.6%, ROC-AUC 0.836). Uni-FinLLM delivers the best overall performance, reaching 84.1% accuracy, an F1-score of 83.6%, ROC-AUC of 0.892, and PR-AUC of 0.871, demonstrating superior robustness under class imbalance. These gains suggest that multimodal fusion of firm fundamentals, structured credit indicators, and financial text enables Uni-FinLLM to better separate high-risk from low-risk entities and capture early warning patterns that single-modality models often miss.

**Table 2.** Credit-Risk Prediction Performance

| Model | Accuracy | F1-Score | ROC-AUC | PR-AUC |
|---|---|---|---|---|
| Logistic Regression | 71.3% | 69.1% | 0.742 | 0.701 |
| Random Forest | 74.5% | 73.4% | 0.781 | 0.731 |
| FinBERT | 77.2% | 76.8% | 0.812 | 0.784 |
| Llama-Fin | 79.6% | 78.4% | 0.836 | 0.813 |
| **Uni-FinLLM (ours)** | **84.1%** | **83.6%** | **0.892** | **0.871** |

Table 3 presents the macro systemic-risk early-warning performance across econometric, deep learning, and graph-based baselines, showing clear benefits of the unified multimodal framework. Models using limited systemic indicators (e.g., Logistic Regression with SRISK inputs) provide a reasonable baseline (Accuracy 68.5%, F1 64.1%, ROC-AUC 0.743), while a VAR model yields only marginal improvement (Accuracy 70.4%, ROC-AUC 0.762). Neural time-series modeling (LSTM-Macro) improves crisis identification (Accuracy 72.8%, F1 68.7%), and the graph-based GNN-MacroRisk captures contagion effects more effectively (Accuracy 75.6%, F1 72.4%, ROC-AUC 0.816). Uni-FinLLM achieves the strongest results with 82.3% accuracy, a crisis F1-score of 79.8%, and ROC-AUC of 0.873, indicating substantial improvement in both overall detection and crisis-period sensitivity. This suggests that jointly modeling macro indicators, market-wide stress signals, and cross-entity relations within a shared representation space enhances early-warning reliability for regulators and stability-monitoring institutions.

Table 3. Macro Systemic-Risk Early-Warning Performance

| Model | Accuracy | F1-Score | ROC-AUC |
|---|---|---|---|
| Logistic Regression (SRISK inputs) | 68.5% | 64.1% | 0.743 |
| VAR Model | 70.4% | 66.5% | 0.762 |
| LSTM-Macro | 72.8% | 68.7% | 0.789 |
| GNN-MacroRisk | 75.6% | 72.4% | 0.816 |
| **Uni-FinLLM (ours)** | **823%** | **79.8%** | **0.873** |

*4.4 Discussion*

The results demonstrate that Uni-FinLLM consistently outperforms both classical econometric methods and domain-specific transformer baselines across micro-level, meso-level, and macro-level financial tasks. The substantial improvement in micro-level directional accuracy—from 61.7% with Llama-Fin to 67.4% with our model—highlights the benefits of multimodal fusion, particularly the integration of structured market time-series with textual sentiment cues. The performance gain in credit-risk prediction, where Uni-FinLLM achieves an ROC-AUC of 0.892, further confirms its strength in handling heterogeneous features such as firm fundamentals, credit indicators, and regulatory disclosures. Most importantly, the model exhibits strong systemic-risk forecasting ability, improving early-warning accuracy to 82.3%, indicating its suitability for central banks, regulators, and macro-prudential surveillance institutions. We attribute these gains to the unified architecture, which enables cross-modal learning and shared representation transfer between micro-level and macro-level tasks, allowing the model to internalize both asset-specific signals and economy-wide risk patterns. Overall, the experimental results validate Uni-FinLLM as a robust decision engine capable of supporting institutional and national-level financial stability assessments.

## 5. Conclusion

This study addresses the challenge of leveraging heterogeneous financial signals for both micro-level stock prediction and macro-level systemic risk assessment. Prior approaches often focus on a single task or a single modality (e.g., time series or text), limiting their ability to capture cross-modal and cross-scale patterns. We propose Uni-FinLLM, a unified multimodal large language model with modular task-specific heads, to jointly model market time-series, firm-level fundamentals, and financial text (and other structured risk indicators) for multi-task forecasting. The primary objective is to improve predictive performance and robustness across micro and macro financial tasks within a single, extensible framework.

Empirical results across multiple benchmarks show that Uni-FinLLM consistently outperforms classical machine-learning baselines, finance-adapted language models, and deep learning models. The model achieves strong gains on micro-level stock prediction as well as on credit-risk prediction, while also delivering improved crisis early-warning capability for systemic risk monitoring. Overall, the results suggest that multimodal fusion and shared representation learning provide a meaningful advantage over single-modality or single-task approaches.

These findings have important implications for financial prediction and risk management. First, a unified multimodal architecture can reuse representations across tasks, reducing duplicated modeling efforts while improving generalization. Second, modular task heads enable practical deployment in real-world settings where different stakeholders (e.g., investors, banks, and regulators) require different outputs from the same underlying system.

Finally, improved macro early-warning performance indicates potential value for systemic stability surveillance, offering more reliable signals that could support timely intervention.

This study has limitations. Model performance may depend on data coverage and the quality of labels for rare events (e.g., crisis periods), and the framework may face challenges under distribution shifts across markets and time. Future work could explore (i) stronger interpretability and attribution across modalities, (ii) more rigorous out-of-sample validation across regions and regimes, and (iii) extending the framework to additional tasks such as stress testing, scenario generation, and portfolio-level optimization.

In conclusion, Uni-FinLLM demonstrates that a unified multimodal LLM with modular heads can effectively bridge micro-level stock prediction and macro-level systemic risk assessment. By integrating diverse financial signals in a shared representation space, the framework improves accuracy and robustness, providing a scalable foundation for multi-task financial intelligence applications.